\documentclass[runningheads]{llncs}
\usepackage[utf8]{inputenc}
\usepackage[T1]{fontenc}
\usepackage{amsmath}
\usepackage{amssymb}
\usepackage{graphicx}

\makeatletter
\newcommand{\@chapapp}{\relax}%
\makeatother

\usepackage[title]{appendix}
\usepackage[thmmarks, amsmath, thref]{ntheorem}
\renewenvironment{proof}{{\emph{Proof.}}}{\begin{flushright}\qedsymbol\end{flushright}}
\renewcommand{\qedsymbol}{$\blacksquare$}

\newcommand{\agent}{\mathop{d}}

\newcommand{\Hq}{\mathbf{H}}

\usepackage{dsfont}

\newcommand{\ind}[1]{\mathds{1}\left[#1\right]}

\newcommand\norm[1]{\left\lVert#1\right\rVert}
\newcommand\V[1]{\mathrm{\textbf{var}}\mleft[#1\mright]}

\usepackage{mleftright}
\makeatletter
\newcommand*{\eval}{%
	\def\E@sub{}%
	\def\E@sup{}%
	\E@scripts
}
\newcommand*{\E@scripts}{%
	\@ifnextchar_\E@subscript{%
		\@ifnextchar^\E@supscript\E@finish
	}%
}
\def\E@subscript_#1{%
	\ifx\E@sub\@empty
	\def\E@sub{#1}%
	\else
	\errmessage{E already has a subscript}%
	\fi
	\E@scripts
}
\def\E@supscript^#1{%
	\ifx\E@sup\@empty
	\def\E@sup{#1}%
	\else
	\errmessage{E already has a superscript}%
	\fi
	\E@scripts
}
\newcommand*{\E@finish}[1]{%
	\mathbb{E}%
	\ifx\E@sub\@empty\else _{\E@sub}\fi
	\ifx\E@sup\@empty\else ^{\E@sup}\fi
	\mleft[#1\mright]%
}
\makeatother

\usepackage[labelformat=simple]{subcaption}

\DeclareCaptionLabelFormat{subcaptionlabel}{\normalfont(\textbf{#2}\normalfont)}
\begin{document}
\title{Optimizing Delegation Between Human and AI Collaborative Agents\thanks{This work was partially supported by the CHIST-ERA-19-XAI010 SAI project. M. Conti’s and A. Passarella's work was partly funded by the PNRR - M4C2 - Investimento 1.3, Partenariato Esteso PE00000013 - "FAIR" funded by the European Commission under the NextGeneration EU programme.}}
%
%
\author{Andrew Fuchs\inst{1,2}\orcidID{0000-0001-7191-8781}\and Andrea Passarella \inst{2}\orcidID{0000-0002-1694-612X} \and Marco Conti\inst{2}\orcidID{0000-0003-4097-4064}}
\authorrunning{A. Fuchs et al.}
%
\institute{Department of Computer Science, Universit\`{a} di Pisa \and
Institute for Informatics and Telematics (IIT), National Research Council (CNR)
\email{andrew.fuchs@phd.unipi.it}}
\maketitle              
\begin{abstract}
In the context of humans operating with artificial or autonomous agents in a hybrid team, it is essential to accurately identify when to authorize those team members to perform actions. Given past examples where humans and autonomous systems can either succeed or fail at tasks, we seek to train a delegating manager agent to make delegation decisions with respect to these potential performance deficiencies. Additionally, we cannot always expect the various agents to operate within the same underlying model of the environment. It is possible to encounter cases where the actions and transitions would vary between agents. Therefore, our framework provides a manager model which learns through observations of team performance without restricting agents to matching dynamics. Our results show our manager learns to perform delegation decisions with teams of agents operating under differing representations of the environment, significantly outperforming alternative methods to manage the team.

\keywords{Reinforcement Learning \and Markov Decision Process \and Delegation \and Learning to defer \and Hybrid Decision-making}
\end{abstract}

\section{Introduction}

Assuming a context with humans working directly in collaboration with autonomous/artificial agents, it is essential to enable a team dynamic eliciting the best combined performance or reduce agent-specific costs. For example, think of an autonomous car, where a human driver or an AI agent can take decisions on the next driving action. It is well known \cite{farsData,rahman2019did} that neither the human nor the agent is always the best choice, as either can make mistakes depending on the driving context. Therefore, it is of utmost importance to design a \emph{delegation policy} deciding, at any point in time, who between the human driver and the AI agent should operate the car. More specifically, a reasonable goal is to identify how and when to enable agent actions to maximize overall team performance while also considering any costs when an agent is operating. Of course, the definition of performance will vary with each application, but the need for performance will not. As such, we investigate a method for guiding delegation decisions from the perspective of a managing authority which selects the agent who chooses the action(s) the team will take in their shared current state. 

With the emergence of a hybrid team comprised of human and artificial agents, it is essential we provide oversight or imbue the agents with an ability to identify the key performer in a given context. To this end, we define and train the model for a Reinforcement Learning (RL) based manager which identifies desirable delegation decisions given context and knowledge regarding agent performance. To ensure a more realistic team dynamic, we assume the manager is not directly observing the individual actions of the agents, but rather that the manager must learn to delegate strictly by learning an association between agents, states, delegations, and outcomes.

In this paradigm, such a manager will allow for behavior learning through indirect observations of agents, with those observations being without any unfair access to private or domain-specific knowledge. In addition, this approach reduces dependencies between the model through which the \emph{agents} learn how to behave, and that through which the \emph{manager} learns how to delegate. With this approach, we will show that our manager can learn a suitable policy. Further, the results demonstrate how the learned policies properly align with the reward function, and how the manager's performance exceeds that of relevant reference alternatives.

In the paper we define a learning model for the manager that can be cast to a well-formulated Reinforcement Learning instance based on a Markov Decision Problem (MDP, see Section~\ref{sec:background}). Without loss of generality, we assume that both the human and the AI agent can be modeled through their own MDP, whose specific actions and policies for transition are independent from each other, and not under the control of the manager. The manager is thus modeled through a separate MDP, whose policy we show how to optimize, based on the underlying policies of the individual agents. This allows us to analytically define optimal policies for the manager, that can be learned according to standard RL machinery. This is quite important, as we demonstrate the possibility of formal optimization of the problem addressed.

In addition, we highlight two further contributions of this work, which distinguish our work from prior literature (analyzed in Section~\ref{sec:related_work}). First, our inclusion of a manager delegating between agents moving through the \emph{same} state space with \emph{different} actions and transitions is a novel conceptualization. The closest alignment to this view is the use of different options (defined in Section~\ref{sec:background}) for moving inside a unique MDP, and their corresponding option policies. In this case, a unique MDP is split into several sub-problems which are managed through a separate learning problem. However, this does not consider managing \emph{multiple} MDPs for solving a given problem, as in our case, where each MDP represents a specific managed agent, as explained in Section~\ref{sec:models}.

Another clear distinction is our consideration of distinct and single agent actions at a given time. This means we are not allowing concurrent actions for the various agents of the team. Rather, the task hinges on the manager's ability to learn an association between contexts and individual agent performance. Therefore, the concepts in the multi-agent cases do not apply.

To summarize, the allowance of distinct and compatible models of the individual agents' MDPs is a worthwhile and novel concept. Under this paradigm, we can remove the common assumptions of many domains while allowing for compatibility with prior RL and related theory. These methods allow us to show analytical results as well as test agent performance.

Before presenting our approach to model and optimize the manager (Section~\ref{sec:models}) and the performance achieved in a relevant case (Section~\ref{sec:evaluation}), in Section~\ref{sec:background} we summarize the main background we exploit in terms of analytical approach, and in Section~\ref{sec:related_work} we analyze in more details the relations and advancements with respect to the state of the art.

\section{Background}
\label{sec:background}

\subsection{Markov Decision Process (MDP)}

Given the use of RL agents to define both our underlying agents and the manager, we assume that the acting agents, those performing actions in the environment, operate with respect to a Markov Decision Process (MDP). We assume a standard definition of the MDP $M = \langle S, A, R, T, \gamma\rangle$, where:
\begin{itemize}
    \item $S$, the state space, is the set of states the agent can traverse
    \item $A$, the action space, is the set of actions an agent is allowed to take
    \item $R: S\times A\rightarrow\mathbb{R}$, is the reward function denoting action utility
    \item $T: S\times A\times S\rightarrow[0, 1]$, the transition function, denotes the probability of transitioning between two states when an action is performed
    \item $\gamma$, the discount parameter, dictates how strongly values decay according to their temporal distance
\end{itemize}
With the assumption of an MDP, we know the actions of the agents depend specifically on the current state/context.

\subsection{Semi-Markov Decision Process (SMDP)}\label{sec:options}

For our manager, we encounter parallels with concepts investigated with regards to Semi-Markov Decision Processes (SMDP), which allows for splitting the problem into a hierarchy of representations. The base level includes a model of discrete states and actions, with the addition of a policy which guides the action selection at this bottom level. Higher levels abstract state and/or action representations into options. The use of options enables treating a sequence of states and actions as a singular item at the higher level of abstraction, which enables behavior learning to at an abstracted level and at the lower level.

In the case of a SMDP, the Markov assumption is relaxed. In one sense, this can be a result of temporally extended actions at the lower abstraction level, where the transitions depend on more than just current state and the selected action. In the other case, SMDPs allow actions conditioned on the history of states and actions. This history is commonly a subset of the episode history and only includes information since the start of a particular time window. The window depends on the use case, but a common one is that of options in which a hierarchy can be formed with a collection of actions treated as a single abstract action (i.e., an option). In this case, the option could take multiple time steps to complete, with the total time being stochastic. As such, the estimates of transitions and rewards need to accommodate the variance in times and actions. A common representation is options $\langle \mathcal{I}, \pi, \beta\rangle$ where:
\begin{itemize}
    \item $\mathcal{I}\subseteq S$ is the set of states where option $o$ is available for selection/activation
    \item $\pi: S\times A\rightarrow[0, 1]$ is the policy for behavior used for the option
    \item $\beta: S^+\rightarrow[0,1]$ is the likelihood an option terminates in a state
\end{itemize}
with $S^+$ referring to the original state space with an included optional terminating/trap state. As will be exemplified in Section~\ref{sec:fixed_window_lens}, such a trap state allows to easily model terminating conditions once an agent has reached a certain goal and should stop any further actions afterwards.

In the SMDP model, the notion of rewards and transitions are updated to consider the new form of interaction agents have with the environment. Rewards are now defined as
\begin{equation}
    r^o_s = \eval{r_{t+1} + \gamma r_{t+2} + \cdots + \gamma^{k-1}r_{t+k}|\mathcal{E}(o,s,t)}
\end{equation}
where $t+k$ represents the random option termination times according to $\pi$ and $\beta$ and $\mathcal{E}(o,s,t)$ is an indicator denoting option $o$ was initiated in state $s$ at time $t$. In this representation, we see the length of an option and the observed rewards dictating the overall option's reward. This accounts for both the stochastic time and the behavior policy underlying agent interaction with the environment. As for state predictions, the common form is a decaying sum of joint probabilities
\begin{equation}
    p^o_{ss'} = \sum^\infty_{k=1}p(s',k)\gamma^k
\end{equation}
where $p(s',k)$ is the probability that the option terminates in state $s'$ after $k$ time steps. Note that this does not define a valid distribution but ensures estimates at longer horizons offer much less importance than the more accurate and shorter horizon estimates. These rewards and visitation estimates can then be combined to define a value function
\begin{equation}
    V^\mu(s) = \eval{r_{t+1} + \gamma r_{t+2} + \cdots + \gamma^{k-1}r_{t+k} + \gamma^kV^\mu(s_{t+k})|\mathcal{E}(\mu,s,t)}
\end{equation}
for Markov policy $\mu$. The policy $\mu$ selects the option to use in state $s$ while $k$ is the option duration. The state-action value can be estimated similarly. Further, this enables temporal difference methods for policy learning. Specifically, we can define a standard $Q()$ function of a Reinforcement Learning problem \cite{sutton2018reinforcement}
\begin{equation}
    Q(s,o) \leftarrow Q(s,o) + \alpha\left[r + \gamma^k\max_{o'\in O_{s'}}Q(s',o') - Q(s,o)\right]
\end{equation}
where the use of $\gamma^k$ to discount the future value follows \cite{bradtke1994reinforcement}.

\section{Related Work}\label{sec:related_work}

\subsection{Hybrid Reinforcement Learning for Single Agent}

In \cite{garcia2020learning}, the goal is to define a model and training method for option learning in a single agent setting. More specifically, the approach focuses on learned options which are ``reusable'', where agents are only expected to add new options if they improve the performance. Another relevant topic is that of \cite{erskine2022developing}, where the authors generate Cooperative Consecutive Policies (CCP). This is intended to generate multiple policies which can be utilized sequentially to accomplish a larger task. The goal with CCP is to reduce the complexity of any one task while improving the compatibility/performance with respect to the consecutive policies.

Generally, approaches found in the single-agent Hierarchical Reinforcement Learning (HRL) literature are more closely aligned to our approach but are attempting to learn policies at both levels of the hierarchy. Instead, we assume our option policies are controlled by existing behavior policies for our available delegation agents. This means the manager agent is not learning the option policies but is instead learning a delegation policy to select ``options'', which in our case are represented by our agents. This means we can instead focus on more direct behavior learning for the manager, but we will need to still ensure that the value function reflects the dynamics the manager will observe when selecting between existing agents.

\subsection{Hybrid Reinforcement Learning in Multi-Agent Reinforcement Learning Contexts}

There are numerous approaches attempting to support multiple agents in a Reinforcement Learning setting, including cases regarding the representation of agent behaviors via ``options'' in fully or partially observable environments \cite{chakravorty2019option,chen2022multi,kurzer2018decentralized,lau2012coordination,menda2018deep,rohanimanesh2002learning,singh2020hierarchical,yang2022ldsa}. Similar to these approaches of temporal abstraction, \cite{guan2022hierarchical} demonstrates state abstraction in a pursuit-evasion game. As a relevant multi-agent HRL example, consider the approach outlined in \cite{yang2022ldsa} which learns to optimally assign agents to sub-tasks in a cooperative multi-agent task. In \cite{yang2022ldsa}, similar agents are grouped so they can share knowledge and be selected for compatible tasks. Selection is performed based on agent ability regarding the task. In this case, the ability measure compares agent experience with the sub-task to compare the relation of its knowledge to the features of the proposed task. This gives a likelihood of a sub-task being selected for an agent which is based on a notion of compatibility.

As we are in fact utilizing multiple agents, it would seem at first like there could be significant relevance to our scenario. In our case, as we are not including cases of simultaneous/concurrent control, we focus on what could equate to a single agent HRL scenario. The manager is only responsible for identifying an agent to delegate and then observing the resulting state changes. Therefore, the manager is learning a policy more closely aligned to the single agent version of this topic. We could conceivably extend our approach in the future to include more agents, but the use of single agent control seems more immediately relevant and far less explored.

\subsection{Delegation}

Previous work in delegation, such as \cite{9821063,s23073409,jacq2022lazy,balazadeh2020switch,straitouri2021triage}, demonstrate relevant concepts and approaches. However, it is worth noting that all of these are assuming the traditional use of single MDP models for all agents. This means that generally the actions, transitions, etc. are consistent for the different agents. In other words, the decision to change delegation comes down strictly to a measure of performance/cost per time step within a common model. This means the variation will come from a difference of agent cost, policy skill levels, observation abilities, etc. In this case, the delegation is occurring strictly between agents operating in entirely equivalent situations but with variations in agent performance. Again, this means that the current common methods focus on a measure of cost or agent performance in determining the best choice.

\section{Agents and Manager Models}
\label{sec:models}

\subsection{Team Agent Models}\label{sec:agent_model}

\begin{figure}[ht]
    \vspace{-3mm}
    \centering
    \includegraphics[width=0.45\textwidth]{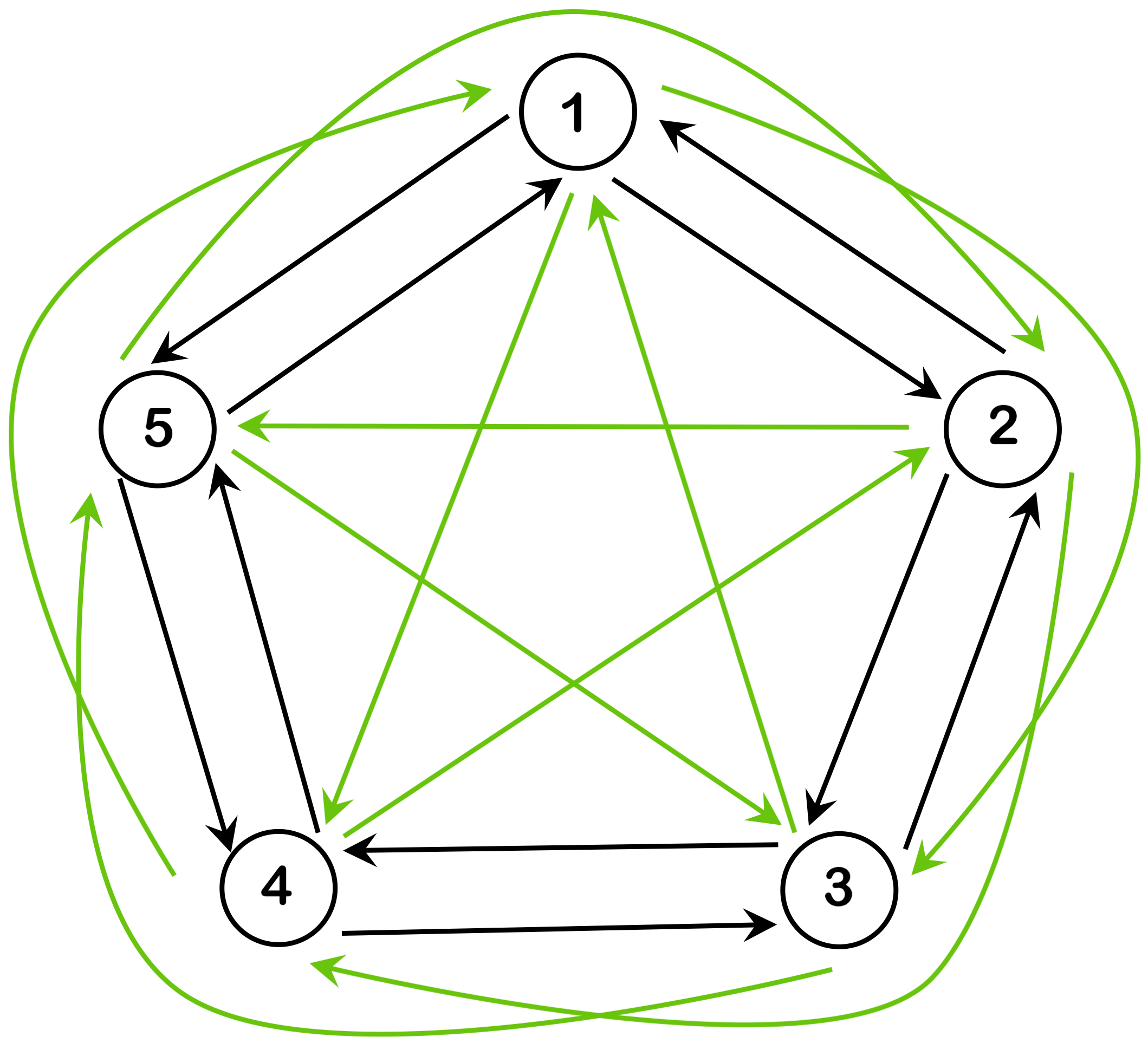}
    \caption{Sample MDPs based on two transition functions (Agent \#1: black arrows, Agent \#2: green arrows)}
    \label{fig:sample_mdp}
    \vspace{-5mm}
\end{figure}

To demonstrate our model and assumptions, we will describe key aspects from the perspective of a team of two underlying agents. Extending our assumption of operating in an MDP, we assume the agents are not required to be operating in the same MDP but rather in compatible MDPs (see Section~\ref{sec:MMDP}). We assume agents exist in the same state space, but our approach allows for differences in actions, transitions, and rewards. As demonstrated in Figure~\ref{fig:sample_mdp}, two agents could exist which operate in the same state space, but transition differently between states. In Figure~\ref{fig:sample_mdp}, these transitions are represented by green arrows for the first agent and black arrows for the second. Both agents can reach the same states (i.e., equivalent state reachability), but the trajectories would vary depending on which agent you observe. This ensures the agents are never given acting authority in a state which does not exist in their MDP, or in a state which could not be reached in their learning phase. Further, this ensures that mixed trajectories utilizing decisions from one or more of the agents remain valid.
The key distinction in our MDP assumptions comes from our allowance of mismatched actions, transitions, and rewards. In this paradigm, the remaining key characteristic is ensuring the transitions maintain the reachability/compatibility assumption. In other words, following the example in Figure~\ref{fig:sample_mdp}, the transitions of each agent only need to ensure that a state reachable by one agent can also be reached by the other. For example, the black arrows allow moving from $1\rightarrow 2$, but the agent following the green arrows would need an additional transition to reach the same state (e.g., $1\rightarrow 4\rightarrow 2$). With the defined requirements on the transitions, we ensure agents are not restricted when learning policies. This ensures we do not restrict the supported agents in our framework, which enables more generalized support for team compositions.

\subsection{Manager MDP \& Model}\label{sec:MMDP}

Following the previous assumption of two agents, we assume there exist two compatible MDPs $M_{d_1} = \langle S_1, A_1, R_1, T_1, \gamma_1\rangle$ and $M_{d_2} = \langle S_2, A_2, R_2, T_2, \gamma_2\rangle$ where $S_1 = S_2$. Further, assume $A_1 \neq A_2 \textrm{ and } T_1 \neq T_2$. Finally, letting $S_M = S_1 = S_2$, our reachability assumption noted in Section~\ref{sec:agent_model} requires that a state $s'\in S$ reachable by agent $\agent_1$ from some state $s_1\in S$ must be reachable from some other state $s_2\in S$ for agent $\agent_2$. Again, this does not require that $s_1 = s_2$, but we do require that each agent can form trajectories that start and terminate in the same states. With these assumptions, it is clear a manager $m$ delegating between agents $\agent_1$ and $\agent_2$ for $M_{d_1}$ and $M_{d_2}$, respectively, can treat the team as navigating a combined Managed MDP (MMDP) $M$. We will define MMDP $M = \langle D, S, A_M, R_M, T_M, \gamma_M\rangle$ with:
\begin{itemize}
    \item $D = \{\agent_1, \agent_2\}$, the set of agents available for delegation
    \item $A_M = \bigcup_{d\in D}\{(a,d): a\in A_d\}$, the action space for MDP $M_i, i\in D$
    \item $R_M(s,d,s')$, the manager's reward function
    \item $T_M$, the manager's transition function
\end{itemize}

With our definition of the MMDP, we can define our relevant manager-specific RL model. As our scenario relates to the options framework from the SMDP model, our manager model is similarly inspired. First, we will define the manager transition, which will utilize the underlying dynamics of the agents in the team:
\begin{equation}
    T_M(s'|s, d) = \sum_{a\in A_d} \pi_d(a|s)T_d\left(s'|s, a\right)
\end{equation}
As is indicated, the likelihood of transitions relies directly on the agent policy $\pi_d$ and transitions $T_d$ for the delegated agent. This allows us to model the single action transitions the manager could encounter given a team. With our transition model, we can then define a value function for our manager given a manager policy $\pi_m$:
\begin{align}\label{eq:manager_V}
    V^{\pi_m}(s) &= \eval_{d\sim\pi_m}{\sum^\infty_{i=0} \gamma^i r_{t+i}|s_t=s, \pi_d}\\\nonumber
                 &= \eval_{d\sim\pi_m}{r_t + \gamma V^{\pi_m}(s_{t+1})|s_t=s, \pi_d}\\\nonumber
                 &= \sum_{d\in D}\pi_m(d|s)\sum_{a\in A_d} \pi_d(a|s)\sum_{s'\in S} T_d(s'|s,a)\left[R_M(s,d,s') + \gamma V^{\pi_m}(s')\right]
\end{align}
Again, the key changes over a standard RL model are the need to account for the policy and transitions of the delegated agents. This gives the definition of the optimal value function $V^*$ as
\begin{equation}
    V^*(s) = \max_d\sum_{a\in A_d}\sum_{s'\in S}\pi_d(a|s)T_d(s'|s,a)\left[R_M(s,d,s') + \gamma V^*(s')\right]
\label{eq:optimal_V}
\end{equation}
with the corresponding state-action value function $Q^*$
\begin{equation}
    Q^*(s,d) = \sum_{a\in A_d}\sum_{s'\in S}\pi_d(a|s)T_d(s'|s,a)\left[R_M(s,d,s') + \gamma V^*(s')\right]
\label{eq:optimal_Q}
\end{equation}

It can be shown that the above definitions ensure the model conforms to the requirements for convergence in Reinforcement Learning (see \ref{appendix:mismatch_q_convergence}). Therefore, we can use standard RL techniques to estimate the optimal manager policy $\pi_m$, i.e., we can optimize the choice of the delegation agent $d$ that maximizes Equation~\ref{eq:manager_V} to maximize the expected reward achieved by the manager (which clearly means optimizing the behavior of the entire system).

\section{Application in Case of Single Action Delegations}\label{sec:fixed_window_lens}

\begin{figure}[ht]
    \vspace{-3mm}
    \centering
    \includegraphics[width=0.45\textwidth]{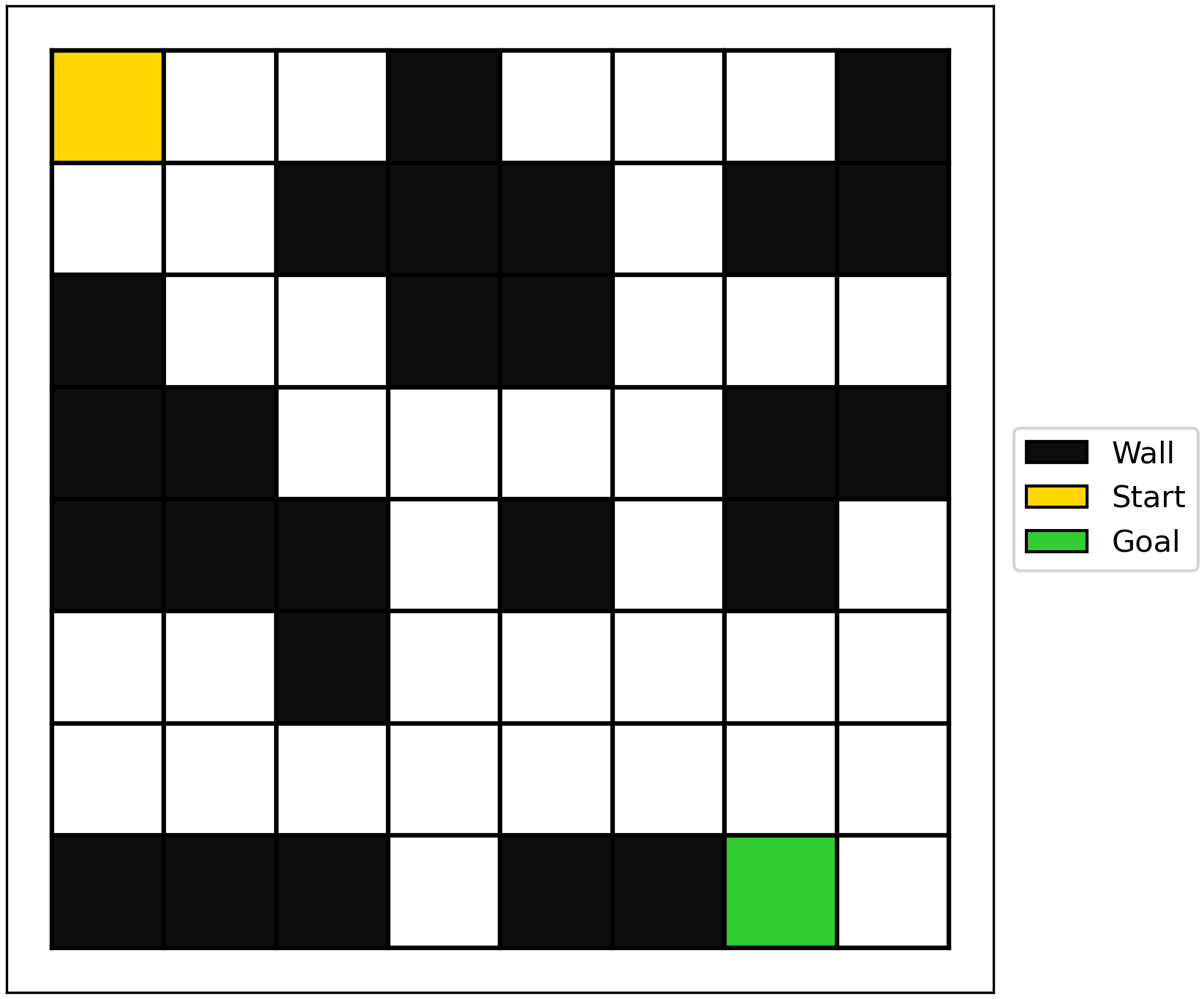}
    \caption{Gridworld environment for training and testing of agents and teams.}
    \label{fig:gridworld}
    \vspace{-5mm}
\end{figure}

To demonstrate our approach, we will test the manager model in a gridworld scenario  (see Figure~\ref{fig:gridworld}). This allows us to test our approach in a scenario which removes possible ambiguous characteristics while still being non-trivial and used in recent literature (e.g., \cite{gabor2019scenario,moy2022evolution}). In a gridworld environment, agents commonly transition through single step actions $a\in\{\uparrow, \rightarrow, \downarrow, \leftarrow\}$. Agents are then trained to accomplish a task, such as safe navigation from a start state to a goal state.

The gridworld also ensures compatible states since agent distinction comes from specific action spaces and transitions for the different agent types. This is done via action spaces $A_i\subset A = \times_k\{\uparrow, \rightarrow, \downarrow, \leftarrow\}$, for agent step size $k$ where we assume that $k$ is a constant parameter, which may be different for each managed agent. In our scenario, the actions for an agent are assumed to take a single time step and are comprised of one or more atomic actions. Therefore, different agents can navigate an environment at different rates, and the manager observes that, when it delegates to a certain agent, that agent will operate for $k$ single step actions in a time step without possibilities for further interventions.
Note, the level of restriction of the action space will determine how likely an agent is to encounter a wall collision. For instance, if we only allow a 3-step agent to move in repeated sequences of atomic actions (e.g., $\rightarrow\rightarrow\rightarrow$), then gridworlds with numerous hallways and turns will increase the frequency of collisions. It is essential the manager identifies where these differences matter and how best to utilize them.

\begin{figure}[ht]
    \vspace{-3mm}
    \begin{subfigure}[t]{0.225\textwidth}
        \centering
        \includegraphics[width=\textwidth]{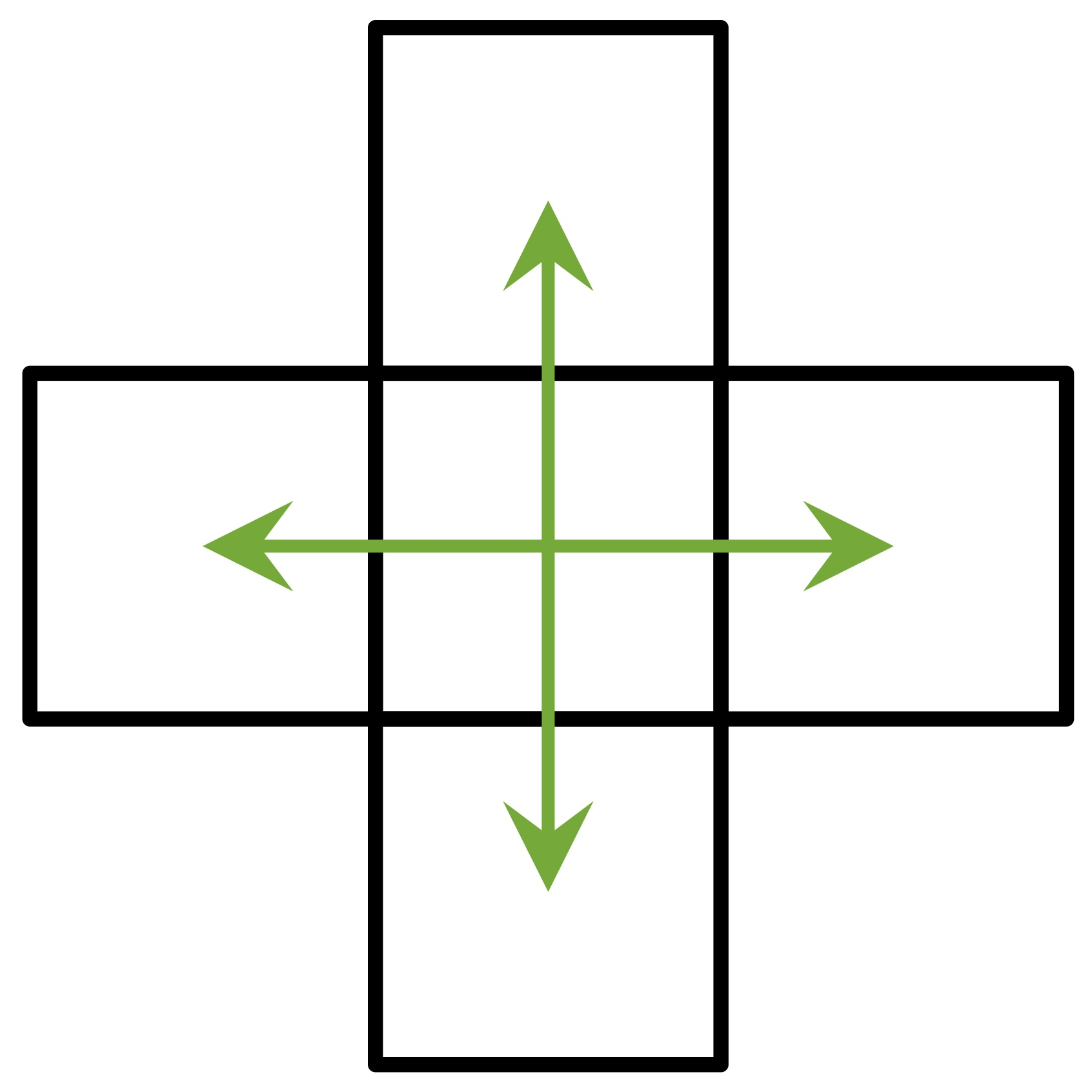}
        \caption{\centering}
        \label{fig:1_step}
    \end{subfigure}
    \hfill
    \begin{subfigure}[t]{0.3\textwidth}
        \centering
        \includegraphics[width=\textwidth]{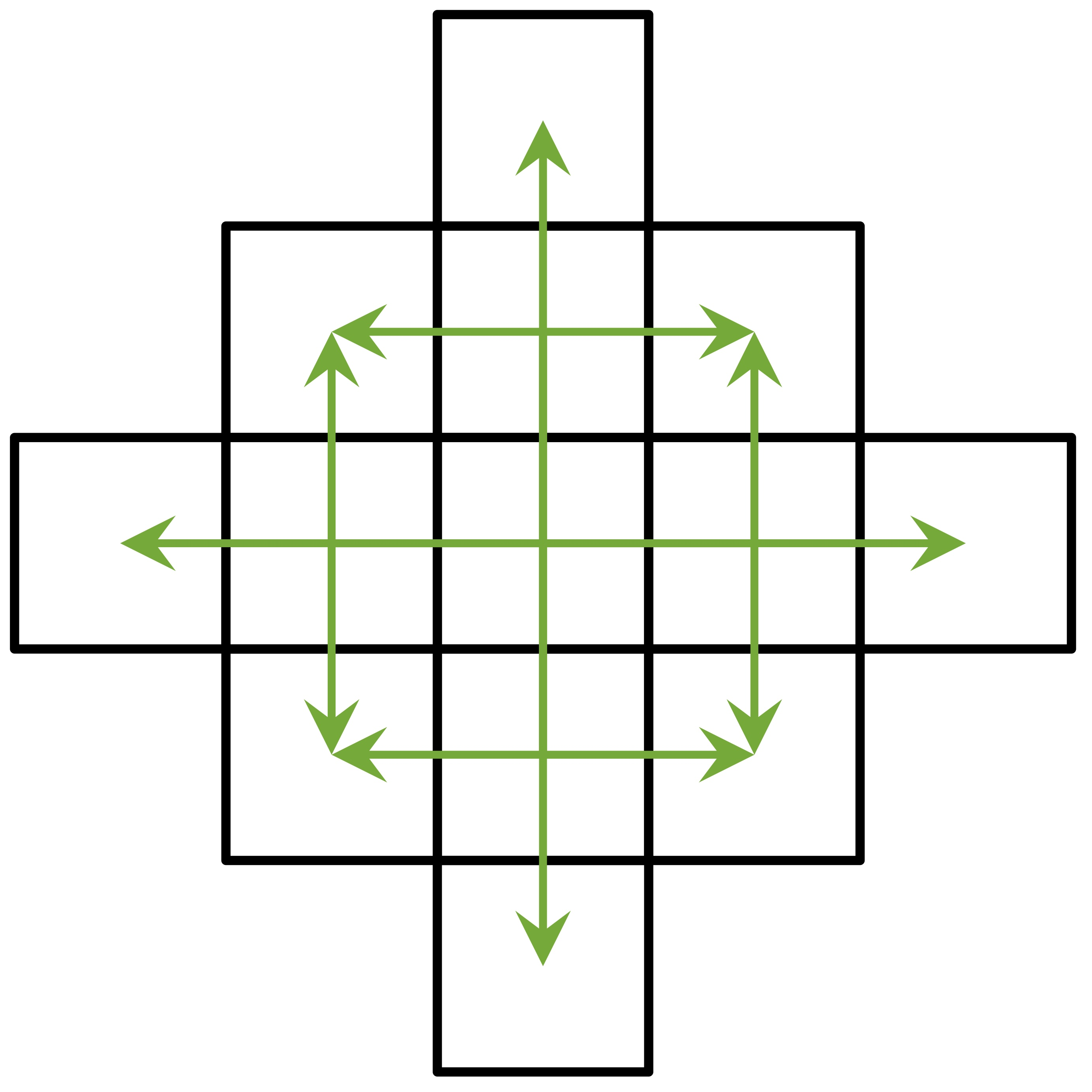}
        \caption{\centering}
        \label{fig:2_step}
    \end{subfigure}
    \hfill
    \begin{subfigure}[t]{0.375\textwidth}
        \centering
        \includegraphics[width=\textwidth]{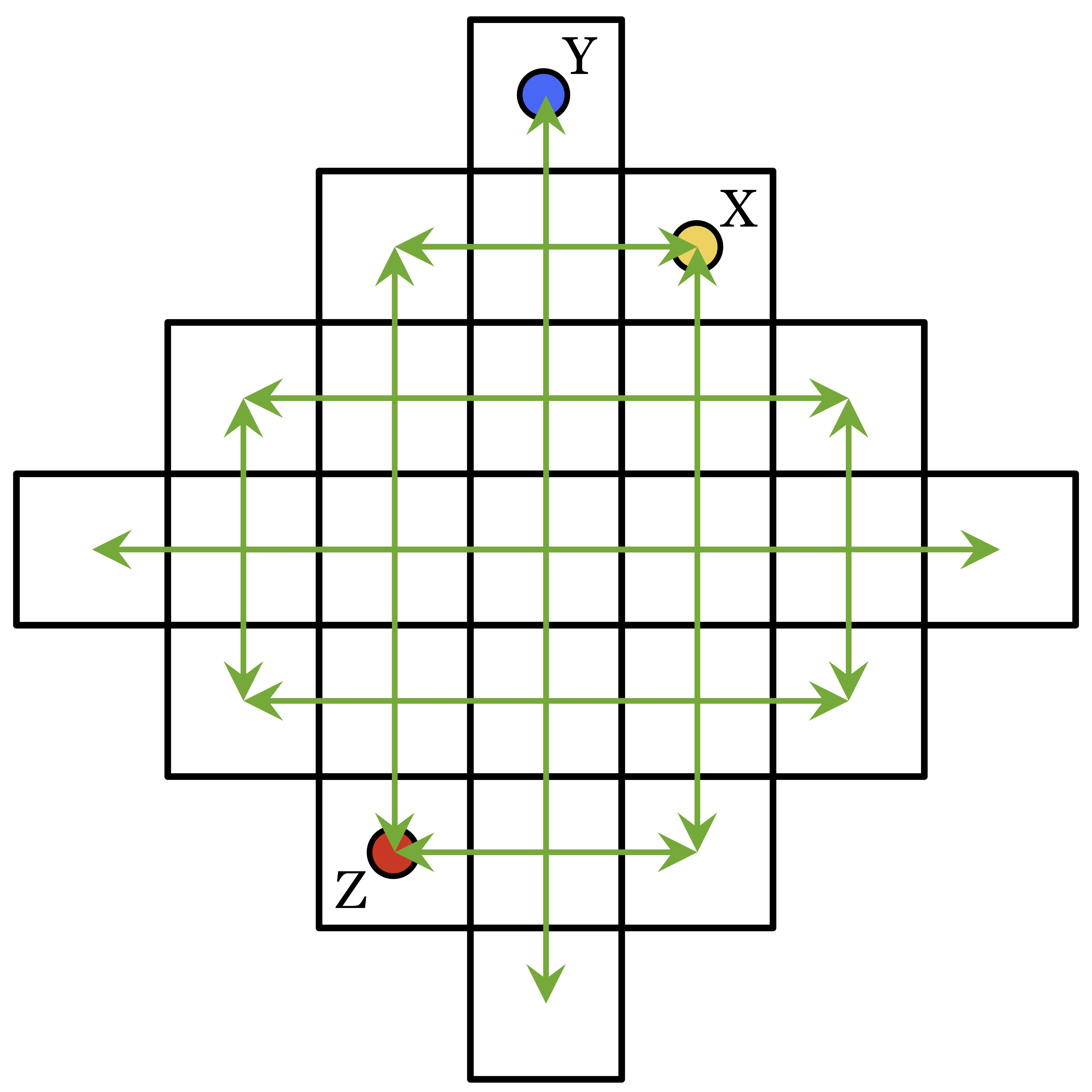}
        \caption{\centering}
        \label{fig:3_step}
    \end{subfigure}
    \caption{Agent actions spaces by step size. Arrows indicate actions starting from center cell. (\textbf{a}) 1-Step agent, (\textbf{b}) 2-Step agent, (\textbf{c}) 3-Step agent with sample action destinations \textbf{X}, \textbf{Y}, and \textbf{Z}.}
    \label{fig:step_actions}
    \vspace{-5mm}
\end{figure}

The actions for our $k$-step agents are demonstrated in Figure~\ref{fig:step_actions}. In the figure, each green arrow indicates an action available to the agent starting from the center point, and the direction/termination of the arrow indicates the movement completed in an unobstructed scenario. Specifically, starting from the center, the agent can terminate in any of the squares containing the end of an arrow, respectively after a 1-, 2- or 3-step action. If the agent encounters a wall along the arrow's path, the action terminates in the cell the agent reached along the arrow's path prior to the collision.

Generally, we can treat any goal state as a trap state with value zero. This will ensure the probability of a trajectory which moves beyond the goal state is zero and that the value of any trajectory with additional steps in the goal state will not accumulate further value. Additionally, a trajectory of step-size $k$ which reaches a goal state after $\hat{k} < k$ atomic actions can be treated as equivalent to any longer sequence with the same sub-sequence. This is due to the probability of all transitions while at the goal state being zero aside from those which loop to stay in the same state with probability equal to one. Hence, early termination of a delegation should have no impact on the manager's transition and value function.

With the above framework, we update our model to include a representation of the $k$-step actions the agents will utilize. This ensures the manager will be observing accurate transitions, which will further ensure the transitions and value function are accurate with respect to the current team.  First, we derive the following manager transition function:
\begin{equation}
    T_M(s'|s,d) = \sum_{a_d(t), s_d(t)}\prod^{t + k - 1}_{i=t}\pi_d(a_i|s_i)T_d(s_{i+1}|s_i,a_i)\ind{s_{t+k} = s'}
\end{equation}
where $a_d(t) =\{a_t,\dots,a_{t + k - 1}\}$ and $s_d(t) =\{s_{t+1},\dots,s_{t+k}|s_t=s\}$ represent the agent's sequence of atomic actions and states during their delegation window. Note that the transition function depends only on the starting and ending states respectively ($s$ and $s'$) and not on the intermediate states, as the latter are not under the control of the manager. This transition function is derived starting from the general formulation of $T_M$ provided in Section~\ref{sec:models}, but we have extended the probability to include the multi-step nature of the agent actions. This ensures the transitions of the manager still reflect the underlying mechanics of the agent MDPs.

Given a definition of manager transitions, we can derive a manager value function. Remember that we assume that the manager has no access to the internal mechanisms through which the managed agents learn their own policies. Therefore, in this context, we will assume that the manager observes no reward derived from the subsequent states and actions of the delegated agents. Therefore, the manager's value function will treat each estimate as independent of these underlying dynamics as well. However, we have included the use of a discount parameter based on the delegated agent's step size. This is inspired by the options framework in SMDPs. In our case, the intermediate rewards that would be discounted in the options framework are unobserved (i.e., are zero) by the manager, so only the $\gamma^{k_d}$ discount is applied to the observed value. As such, the value function will follow a similar pattern to Equation~\ref{eq:manager_V}. In this case, the delegation agent specific actions and transitions are encompassed by $T_M$, resulting in value function.
\begin{equation}
    V(s) = \sum_{d\in D}\pi_M(d|s)\sum_{s'}T_M(s'|s,d)[R_M(s,d) + \gamma^{k_d} V(s')]
\end{equation}
where $k_d$ is the step-size of the delegated agent $d$, while $[R_M(s,d) + \gamma^{k_d} V(s')]$ denotes the expected rewards of the manager when choosing delegation agent $d$ with expected transition in state $s'$. The use of $\gamma^{k_d}$ aligns with the SMDP framework and enables indicating a discount dependent on the number of atomic actions a delegated agent will perform.

For the reward function, we will use a simple reward which does not rely on any knowledge specific to the underlying agents. For our manager, we will provide rewards using
\begin{equation}\label{eqn:manager_reward}
    R=\begin{cases}
        100 - c_d \quad &\, \text{goal is reached} \\
        -1 - c_d \quad &\, \text{goal not reached, but valid step} \\
        -10 - c_d \quad &\, \text{delegated agent caused wall collision} \\
        -100 - c_d \quad &\, \text{episode terminated without reaching goal} \\
    \end{cases}
\end{equation}
where $c_d$ denotes the costs of agent $d$. In other words, we penalize the manager for agent costs, and for episode length with the step penalty in the reward function. Note that the specific values of the reward are not that important. What is key is to have a form of reward that can help drive the manager in the desired direction. With the given reward, the manager should learn to minimize episode lengths while accounting for agent costs. In our paradigm, cost can represent relative preference, operating costs, etc. We should note that this simpler reward does not specifically prevent the manager from allowing a wall collision if the overall reward is higher. In this scenario, we wanted to demonstrate the manager's ability to find a team which can reach the goal fastest and lowest cost, but with the option to accept a collision if it means higher overall reward. Regarding the manager reward, safety-critical cases could be considered. In this case, wall collisions could terminate an episode with a penalty. Alternatively, the penalty for collisions could be increased to further discourage these outcomes.

\section{Results}
\label{sec:evaluation}

In this section, we will demonstrate the performance of our manager in multiple team configurations. Further, our results will include cases where the manager delegations are subject to agent-specific costs. With the inclusion of cost, the manager must make delegations which both match the desire for good team performance while also accounting for agent cost.

The x-axis labels correspond to the various team compositions. Our label convention is "AB-CD", where A/C denote step size and B/D denote error likelihood -- L: low, M: medium, H: high, N: none. For example, 1H-2L corresponds to a team with a 1-step high error likelihood agent and 2-step low error likelihood agent. Likelihood indicates how frequently an agent makes an error and likely severity of change. The actions selected can change with our error mapping via a decaying exponential function (truncated in $[0, \pi]$), which determines how large a change is made. An error occurs when the action performed differs from the originally intended action. The severity of the error indicates how much of a divergence is observed between the error and the intended action.

Following Figure~\ref{fig:step_actions}, each $k$-step action corresponds to an arrow leading from the center cell to another cell. Errors occur when an agent follows a different arrow than intended. As seen in Figure~\ref{fig:2_step} and Figure~\ref{fig:3_step}, different actions/arrows can lead to the same state but different outcomes. For instance, the costs for actions could vary or walls could exist along one arrow's path and not another. In these cases, the rewards or trajectories could vary significantly. As an example of errors, starting from $\uparrow\uparrow\uparrow$ leading to \textbf{Y}, $\uparrow\uparrow\rightarrow$ is a clockwise shift of one arrow/action leading to \textbf{X} and $\leftarrow\leftarrow\downarrow$ is an anticlockwise shift by eight actions/arrows leading to \textbf{Z} (Note: $\downarrow\downarrow\leftarrow$ is a nine-error shift also leading to \textbf{Z}). Errors are determined by sampling according to our function and binning to determine how many arrows away to select the error action, with clockwise/anticlockwise chosen randomly.

Regarding team compositions, we consider cases of agents with step sizes of 1 to 3. This was intended to represent agents with clear distinction in action spaces while maintaining a reasonable level of ``speed'' for the agents (i.e., limiting the advantages of any one agent). Further, the step sizes enable teams with agents demonstrating strong performance in particular grid regions/types, but they also allow for variation between agents. This ensures the manager will have opportunities to operate under multiple team configurations and varying ideal utilization.

\subsection{Low Cost for Shorter Steps}

For the following case, the manager was given a cost of 1, 4, and 7 for the 1-step, 2-step, and 3-step agents, respectively. With this, the manager is given a reward function which counteracts some benefits of choosing agents with higher step sizes. This impact is in conjunction with the fact that the manager reward allows for some wall collisions with penalty. Therefore, we see an increasing per-step cost as the step size increases (e.g., cost of 6 for six 1-step actions vs. a cost of 14 for two 3-step actions). In the following plots, we illustrate the rewards observed for both trained and random managers, i.e., managers that delegate between the two agents according to a uniform probability of 0.5 at any decision point.

\begin{figure}[!ht]
    \vspace{-3mm}
    \begin{subfigure}[t]{0.49\textwidth}
        \centering
        \includegraphics[width=\textwidth]{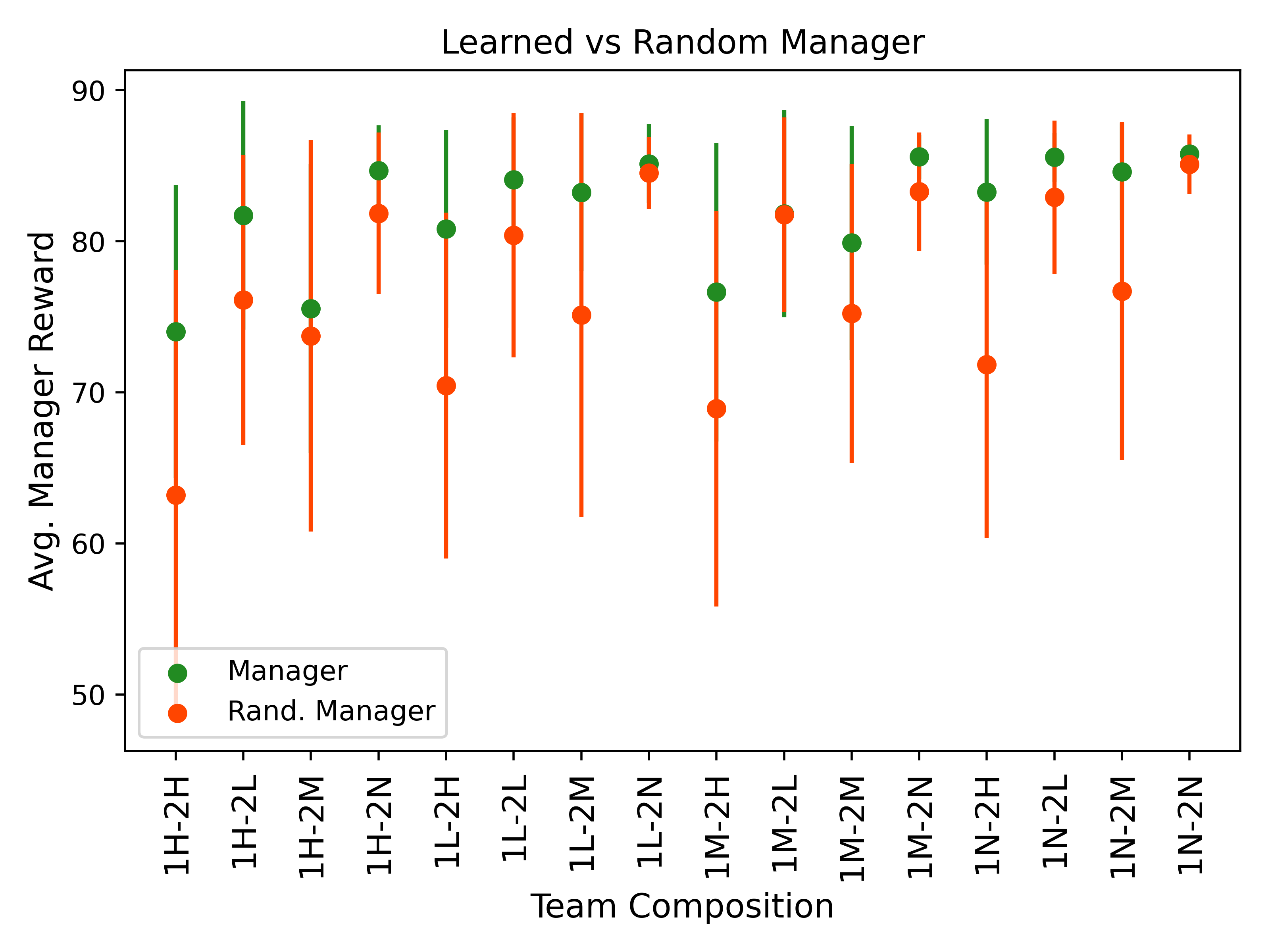}
        \caption{\centering Low-Cost: 1-Step/2-Step team}
        \label{fig:1_4_7_step_1_2_results}
    \end{subfigure}
    \hfill
    \begin{subfigure}[t]{0.49\textwidth}
        \centering
        \includegraphics[width=\textwidth]{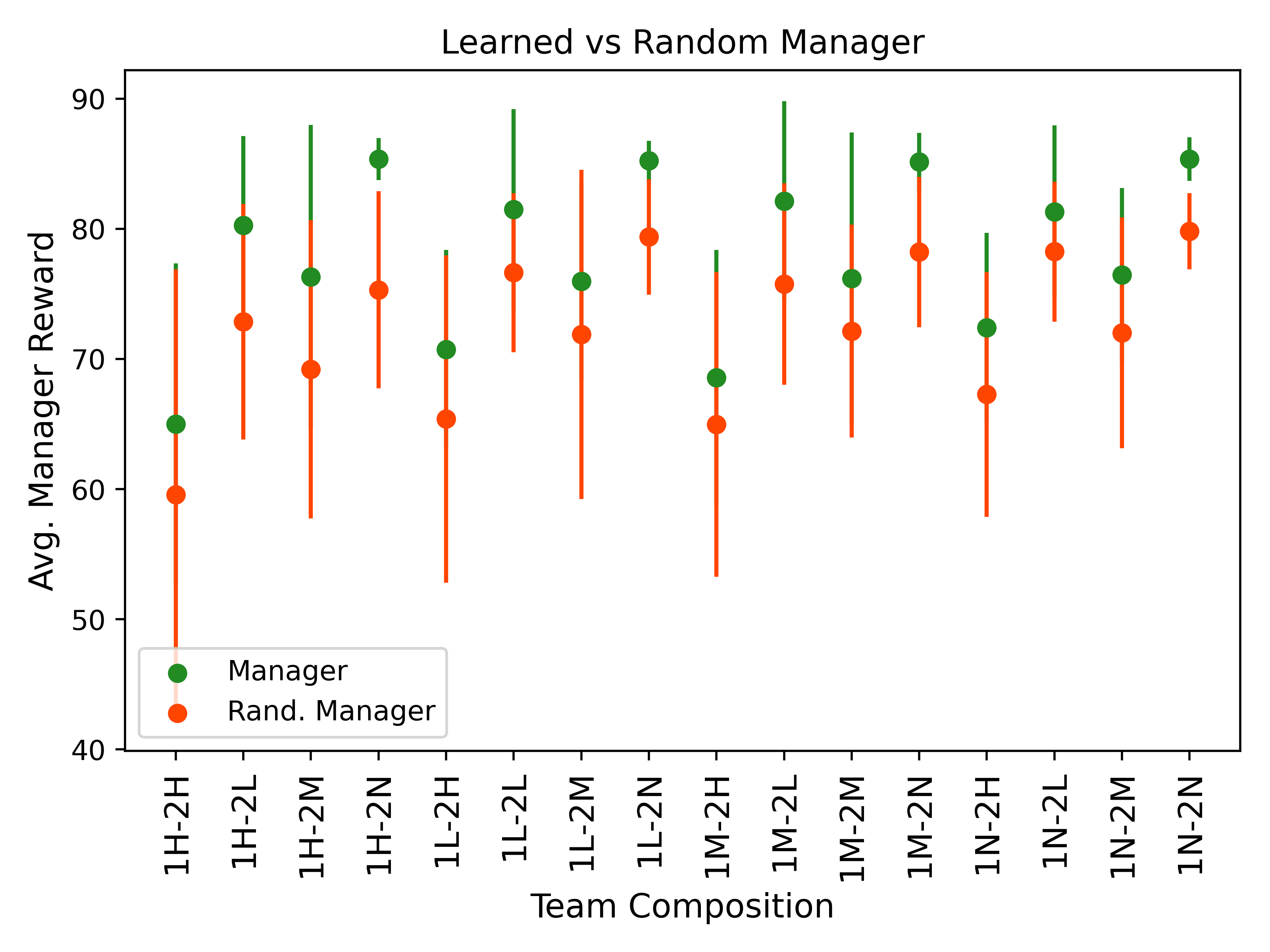}
        \caption{\centering High-Cost: 1-Step/2-Step team}
        \label{fig:7_4_1_step_1_2_results}
    \end{subfigure}
    \hfill\\
    \begin{subfigure}[t]{0.49\textwidth}
        \centering
        \includegraphics[width=\textwidth]{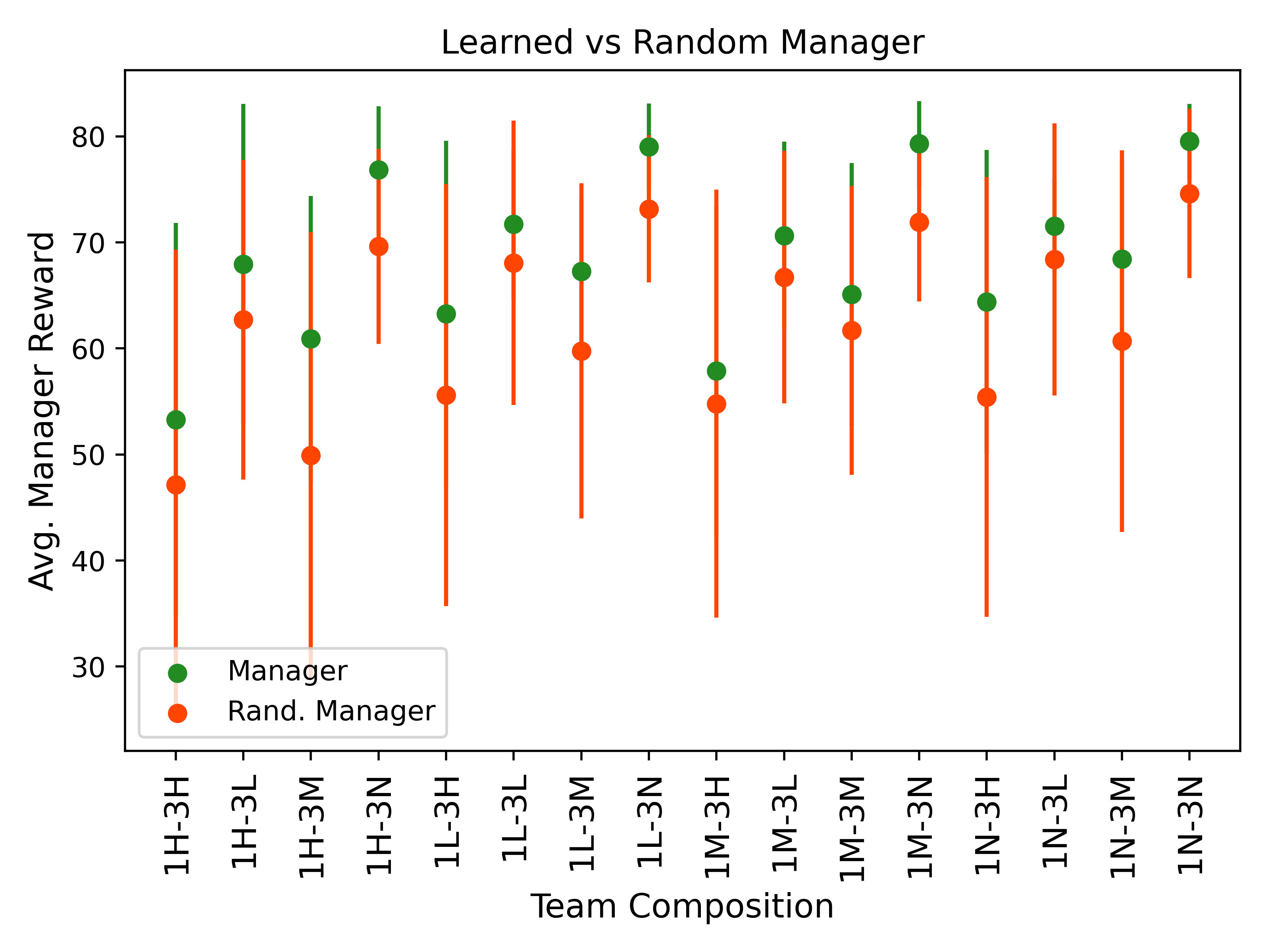}
        \caption{\centering Low-Cost: 1-Step/3-Step team}
        \label{fig:1_4_7_step_1_3_results}
    \end{subfigure}
    \hfill
    \begin{subfigure}[t]{0.49\textwidth}
        \centering
        \includegraphics[width=\textwidth]{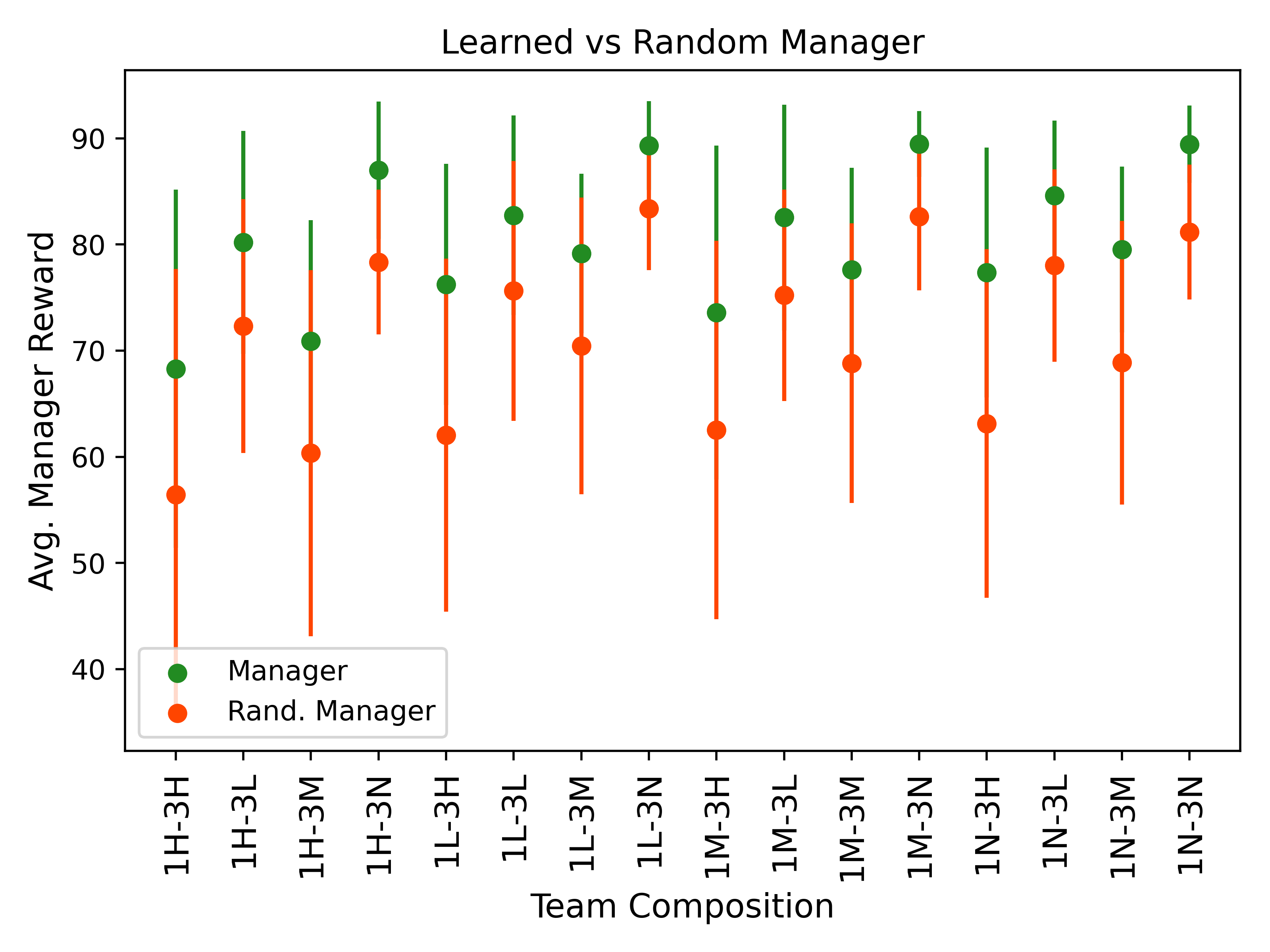}
        \caption{\centering High-Cost: 1-Step/3-Step team}
        \label{fig:7_4_1_step_1_3_results}
    \end{subfigure}
    \hfill\\
    \begin{subfigure}[t]{0.49\textwidth}
        \centering
        \includegraphics[width=\textwidth]{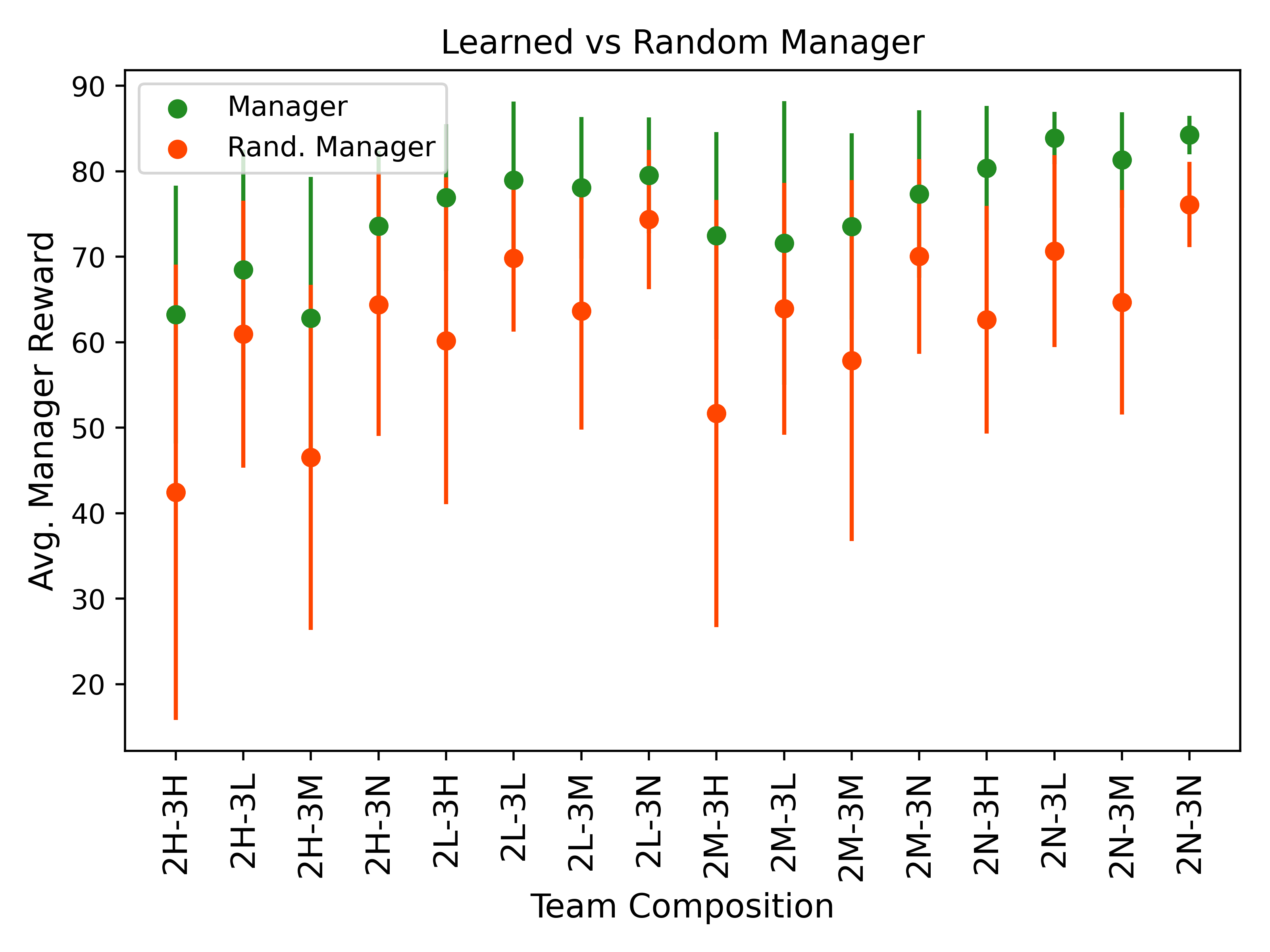}
        \caption{\centering Low-Cost: 2-Step/3-Step team}
        \label{fig:1_4_7_step_2_3_results}
    \end{subfigure}
    \hfill
    \begin{subfigure}[t]{0.49\textwidth}
        \centering
        \includegraphics[width=\textwidth]{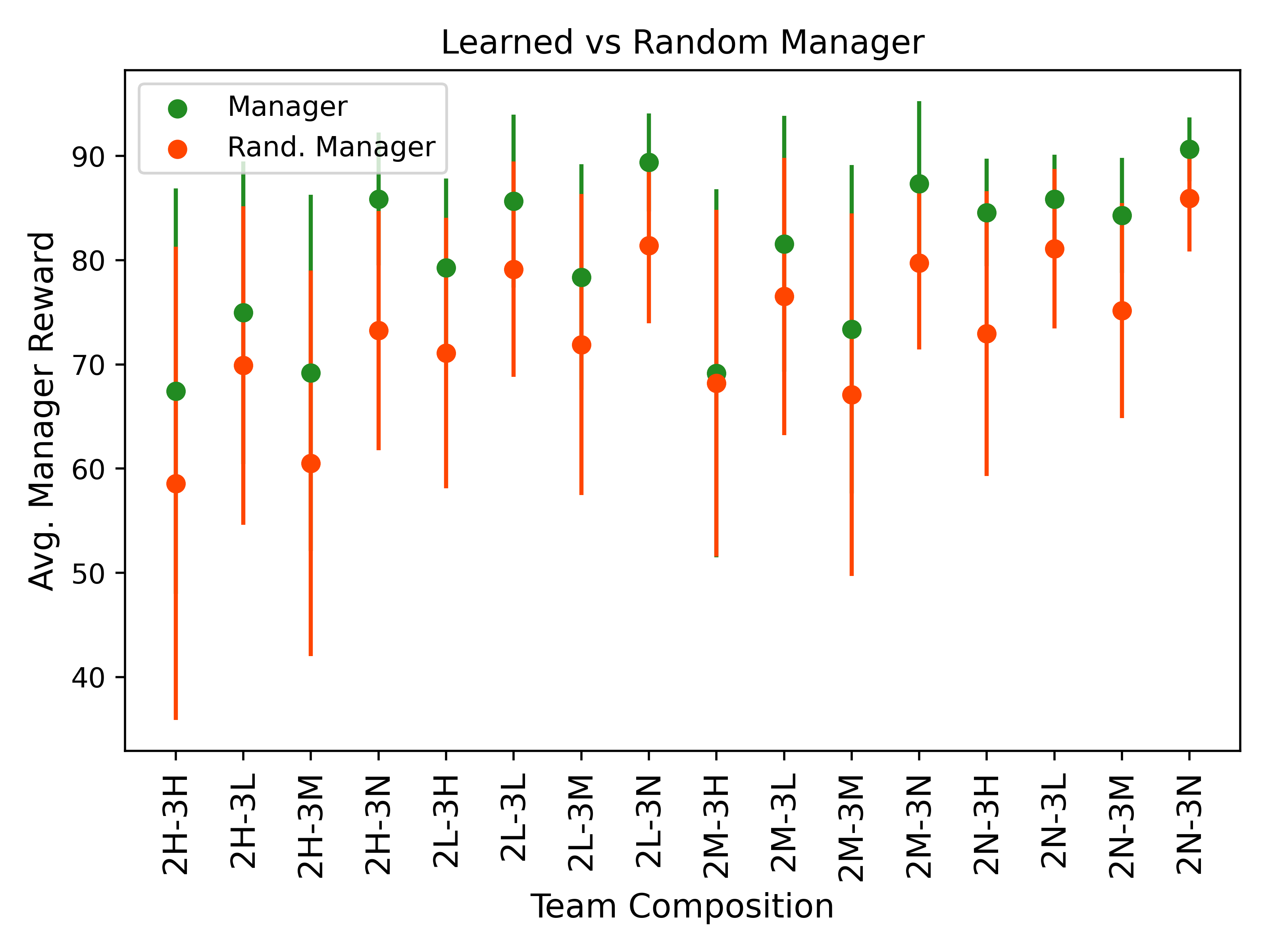}
        \caption{\centering High-Cost: 2-Step/3-Step team}
        \label{fig:7_4_1_step_2_3_results}
    \end{subfigure}
    \hfill
    \caption{Manager performance compared to random selection. Agents subject to varying levels of error likelihood -- L: Low, M: Medium, H: High, N: None}
    \label{fig:results}
    \vspace{-5mm}
\end{figure}

As is apparent in our results, cases with the least error-prone agents offer the manager the best chance of success. For example, the 1N-2N (Figure~\ref{fig:1_4_7_step_1_2_results}), 1L-3N (Figure~\ref{fig:1_4_7_step_1_3_results}), and 2N-3L (Figure~\ref{fig:1_4_7_step_2_3_results}) cases all demonstrate the manager utilizing the low error rates of the agents, resulting in higher overall rewards. Further, the lower error rates allow the manager to take advantage of the higher step sizes to maintain strong team performance. For cases where the manager encounters one or more team members with high error likelihoods, we continue to see cases of strong performance. As seen for 1H-2L (Figure~\ref{fig:1_4_7_step_1_2_results}), 1H-3N (Figure~\ref{fig:1_4_7_step_1_3_results}), and 2L-3H (Figure~\ref{fig:1_4_7_step_2_3_results}), our manager finds agent utilization achieving high scores despite the high variance in agent performances.

\subsection{Higher Cost for Shorter Steps}

In a cost reversal, we demonstrate the case where the manager now observes a much higher cost for the short step agents and reduced for the longer ones. The manager is given agent costs of 7, 4, and 1 for the 1-step, 2-step, and 3-step agents, respectively. While this change should not impact the manager's ability to select better performing agents, we do anticipate seeing modifications to the relative performance and overall manager results (i.e., maximum reward and agent utilization).

We again see our manager learns desirable agent utilization. Clearly, the 1L-2N (Figure~\ref{fig:7_4_1_step_1_2_results}), 1L-3N (Figure~\ref{fig:7_4_1_step_1_3_results}), and 2L-3N (Figure~\ref{fig:7_4_1_step_2_3_results}) cases all demonstrate the manager utilizing low error agents while accounting for their corresponding costs to find optimal results. In the cases of one or more team members with high error likelihoods, we continue to see cases of strong performance. As seen in the 1H-2L (Figure~\ref{fig:7_4_1_step_1_2_results}), 1H-3N (Figure~\ref{fig:7_4_1_step_1_3_results}), and 2M-3M (Figure~\ref{fig:7_4_1_step_2_3_results}) cases, our manager again finds agent utilization achieving high scores despite the high variance in agent performances.

\section{Conclusion}

In Figure~\ref{fig:results}, we see the manager has consistent and stronger performance with less error-prone agents for both agent cost scenarios. In both cases, as expected, increased variance of agent performance coincides with increased variance in manager rewards. Still, given the performance indicated by our method, we see that our manager was successful in learning a delegation policy. Despite the manager operating without direct observation of agent actions, the manager learned identify desirable delegations to generate good trajectories. Consequently, we demonstrated a method which could train a manager to successfully delegate between agents without assuming unfair access to additional knowledge. Further, we demonstrated the manager can accommodate for changes in agent costs to learn a delegation policy.

\paragraph{Acknowledgments}

This work was partially supported by the CHIST-ERA-19-XAI010 SAI project. M. Conti’s and A. Passarella's work was partly funded by the PNRR - M4C2 - Investimento 1.3, Partenariato Esteso PE00000013 - "FAIR" funded by the European Commission under the NextGeneration EU programme.

\bibliographystyle{splncs04}
\bibliography{references}

\section{Appendices}

\begin{appendices}

\renewcommand{\thesection}{Appendix \Alph{section}}

\section[]{Convergence of Q-Learning in Mismatched MDPs}\label{appendix:mismatch_q_convergence}

Following the proof of convergence for Q-Learning on MDPs in \cite{jaakkola1993convergence,melo2001convergence}, we will show that the learning algorithm converges to the optimal $Q^*$.

\begin{proof}
    As with traditional Q-Learning, we will start by showing that the optimal $Q$ is a fixed point of a contraction operator $\Hq$. First, recall that we are considering MDPs $M_{d_1} = \langle S_1, A_1, R_1, T_1, \gamma_1\rangle$ and $M_{d_2} = \langle S_2, A_2, R_2, T_2, \gamma_2\rangle$, with $S_1 = S_2$. Therefore, define
    \begin{equation}
        (\Hq q)(x, d) = \sum_{s'\in S}\sum_{a\in A_d}\pi_d(a|s)T_d(s'|s,a)\left[R_M(s,d,s') + \gamma\max_{d'}q_(s',d') \right]
    \end{equation}
    We show $\Hq$ is a contraction in the sup-norm by showing that $\norm{\Hq q_1 - \Hq q_2}_\infty \leq \gamma\norm{q_1 - q_2}_\infty$. From the definition of $\Hq$, we see that
    \begin{align*}
        \norm{\Hq q_1 - \Hq q_2}_\infty &= \max_{s, d}\left|\sum_{s'\in S}\sum_{a\in A_d}\pi_d(a|s)T_d(s'|s,a)\left[R_M(s,d,s') + \gamma\max_{d'}q_1(s',d') \right.\right.\\ &\qquad\qquad\qquad\qquad\qquad\qquad\qquad\qquad\quad\left.\left. -\, R_M(s,d,s') - \gamma\max_{d'}q_2(s',d')\right]\right|\\
        &= \max_{s, d}\gamma\left|\sum_{s'\in S}\sum_{a\in A_d}\pi_d(a|s)T_d(s'|s,a)\left[\max_{d'}q_1(s',d') - \max_{d'}q_2(s',d')\right]\right|\\
        &\leq \max_{s, d}\gamma\sum_{s'\in S}\sum_{a\in A_d}\pi_d(a|s)T_d(s'|s,a)\left|\max_{d'}q_1(s',d') - \max_{d'}q_2(s',d')\right|\\
        &\leq \max_{s, d}\gamma\sum_{s'\in S}\sum_{a\in A_d}\pi_d(a|s)T_d(s'|s,a)\max_{d'}\left|q_1(s',d') - q_2(s',d')\right|\\
        &= \max_{s, d}\gamma\sum_{s'\in S}\sum_{a\in A_d}\pi_d(a|s)T_d(s'|s,a)\norm{q_1 - q_2}_\infty\\
        &= \gamma\norm{q_1 - q_2}_\infty
    \end{align*}
    Given $\Hq$ is a contraction, we will then show that $Q_t$ converges to $Q^*$ as $t\rightarrow\infty$. To start, we will assume that for $\forall t$, we have $0\leq\alpha_t\leq 1, \sum_t\alpha_t=\infty, \mathrm{ and } \sum_t\alpha_t<\infty$. Let
    \begin{equation*}
        F_t(s,d) = R_M(s,d,X(s,d)) + \gamma\max_{d'}Q_t(X(s,d),d') - Q^*(s,d)
    \end{equation*}
    for next state distribution $X(s,d)$. Then, we will first show that $\norm{\eval{F_t(s,d)|\mathcal{F}_t}}_\infty \leq\gamma\norm{\Delta_t}_\infty$, with $\gamma < 1$. Let
    \begin{equation*}
        \Delta_t = Q_t(s,d) - Q^*(s,d)
    \end{equation*}
    then
    \begin{equation*}
        \Delta_{t+1} = (1 - \alpha_t)\Delta_t + \alpha_tF_t(s,d)
    \end{equation*}
    Therefore,
    \begin{align*}
        \eval{F_t(s,d)|\mathcal{F}_t} &= \sum_{s'\in S}\sum_{a\in A_d}\pi_d(a|s)T_d(s'|s,a)\left[R_M(s,d,s') + \gamma\max_{d'}Q_t(s',d') - Q*(s,d)\right]\\
        &= (\Hq Q_t)(s,d) - Q^*(s,d)\\
        &= (\Hq Q_t)(s,d) - (\Hq Q^*)(s,d)\\
        &\overset{(a)}{\Rightarrow} \norm{\eval{F_t(s,d)|\mathcal{F}_t}}_\infty \leq\gamma\norm{Q_t - Q^*}_\infty\\
        &= \gamma\norm{\Delta_t}_\infty
    \end{align*}
    with $(a)$ resulting from the contraction proof. Next, we will show that $\V{F_t(s)|\mathcal{F}_t} \leq C(1 + \norm{\Delta_t}_\infty)^2$ for $C > 0$.
    \begin{align*}
        \V{F_t(s)|\mathcal{F}_t} &= \eval{\left(R_M(s,d,X(s,d)) + \gamma\max_{d'}Q_t(s',d') - Q^*(s,d) - (\Hq Q_t)(s,d) + Q^*(s,d)\right)^2}\\
        &= \eval{\left(R_M(s,d,X(s,d)) + \gamma\max_{d'}Q_t(s',d') - (\Hq Q_t)(s,d)\right)^2}\\
        &\overset{(a)}{=} \V{R_M(s,d,X(s,d)) + \gamma\max_{d'}Q_t(s',d')|\mathcal{F}_t}\\
        &\leq C(1 + \norm{\Delta_t}_\infty)^2
    \end{align*}
    with the final inequality resulting from \cite{melo2001convergence}. Further, note that $(a)$ comes from the fact that $(\Hq Q_t)(s,d) = \eval{Q_t(s,d)}$. Therefore, we see that $Q^t$ converges to $Q^*$ with probability $1$.
\end{proof}

\end{appendices}

\end{document}